\documentclass[runningheads]{llncs}
\usepackage{multirow}

\usepackage{enumitem}
\usepackage{algorithm}
\usepackage{algpseudocode}
\usepackage{amssymb}
\usepackage{amsmath}

\usepackage[T1]{fontenc}
\usepackage{graphicx}
\begin{document}
\title{Examining the Metrics for Document-Level Claim Extraction in Czech and Slovak}

\titlerunning{Metrics for Document-Level Claim Extraction in Czech and Slovak}
%
\author{
Lucia Makaiová\thanks{Correspondence to: \texttt{makaiovalucia@gmail.com}} \and 
Martin Fajčík \and Antonín Jarolím
}
\authorrunning{L. Makaiová, M. Fajčík, A. Jarolím}
%
\institute{Brno University of Technology, Czech Republic\\
}
\maketitle              
\begin{abstract}
Document-level claim extraction remains an open challenge in the field of fact-checking, and subsequently, methods for evaluating extracted claims have received limited attention. In this work, we explore approaches to aligning two sets of claims pertaining to the same source document and computing their similarity through an alignment score. We investigate techniques to identify the best possible alignment and evaluation method between claim sets, with the aim of providing a reliable evaluation framework. 
Our approach enables comparison between model-extracted and human-annotated claim sets, serving as a metric for assessing the extraction performance of models and also as a possible measure of inter-annotator agreement. We conduct experiments on newly collected dataset---claims extracted from comments under Czech and Slovak news articles---domains that pose additional challenges due to the informal language, strong local context, and subtleties of these closely related languages.
The results draw attention to the limitations of current evaluation approaches when applied to document-level claim extraction and highlight the need for more advanced methods---ones able to correctly capture semantic similarity and evaluate essential claim properties such as atomicity, checkworthiness, and decontextualization.
\keywords{fact-checking  \and claim extraction \and similarity metrics.}
\end{abstract}
\section{Introduction}
\vspace{-0.75em}
Claim extraction is a key stage in the automatic fact-checking pipeline. Its purpose is to identify claims within a document that can be verified during the fact-checking process. 
In practice,  this task is often coupled with the checkworthiness estimation. The resulting claims should therefore be atomic, decontextualized, and check-worthy, which we define as follows. 

\begin{itemize}
\item \textbf{Atomic} claim should express a single factual proposition. We also consider multi-part statements (e.g., ``Marie Curie was a physicist and chemist.'') as atomic if they are presented as one inseparable unit in the source document.
\item \textbf{Decontextualized} claim should be self-contained and interpretable without relying on the surrounding text.
\item \textbf{Check-worthy} claim is factual claim that the general public will be interested in learning about its veracity \cite{arslan2020benchmark}.
\end{itemize}

Our work focuses on the problem of exhaustive document-level claim extraction. We investigate how to evaluate sets of extracted claims in a way that also
properly reflects exhaustivity-ensuring that both under-generation and over-generation (i.e., generating too little, or too many claims respectively) are penalized. To this end, we propose and validate a simple evaluation framework
based on an LLM-as-a-judge metric to assess claim extraction of language models on our collected datasets. They consist of claims extracted from comments under Czech
and Slovak online news articles. Our goal is to find reliable ways to evaluate claim extraction for the purposes of fact-checking. We focus on news comments because they are a space where people actively discuss current topics and frequently spread potentially incorrect or misleading claims. This makes the domain especially important for developing tools that can help verify information and reduce the spread of misinformation. We select Czech and Slovak because these languages are underrepresented in current automated fact-checking research and lack large-scale, high-quality resources for claim detection and evaluation, which makes it challenging to build and improve systems for this task.

Our contributions are:
\vspace{-0.2em}
\begin{enumerate}[label=(\roman*)]
    \item Two novel datasets containing human-annotated document-level claims in Czech and Slovak.
    \item Proposed framework for measuring claim similarity at the document level.
    \item A human preference study to derive more meaningful interpretations of model scores across different evaluation metrics applied within our framework.
\end{enumerate}
\vspace{-1.5em}
\section{Related Work}
\vspace{-0.75em}
\subsection{Similarity Metrics}
\vspace{-0.25em}
\label{ch:similarity}
We evaluate the quality of claim extraction by measuring the similarity between extracted and reference claims. Similarity metrics for comparing textual sequences vary in their approach and robustness. 

Purely lexical methods focus on a surface form of the word, and they fail to capture the claim's similarity to
the reference after paraphrasing \cite{zhang2020bertscoreevaluatingtextgeneration}. They also fail to penalize the change of meaning
properly when the words are kept the same.
For example, if the reference is ``The cat sat on a mat.'' lexical metrics (such as ROUGE or edit distance) typically rate ``The mat sat on a cat.'' higher than ``My cat sat on a rug.'', despite the former being semantically different (and nonsensical). 

Therefore, semantic metrics are prefered, because they can capture the similarity of semantics beyond surface-level lexical cues. BERTScore~\cite{zhang2020bertscoreevaluatingtextgeneration} is an embedding-based metric that computes similarity using contextualized BERT embeddings. Unlike metrics based solely on lexical overlap, BERTScore can capture paraphrasing and contextual differences. This is because the embeddings encode both token identity and surrounding context, causing semantically related words to have a low cosine distance. 

However, embedding-based semantic similarity metrics also have notable weaknesses. One of the main issues is their difficulty in handling words that reverse or substantially change the meaning of a sentence, most prominently negations~\cite{anschutz-etal-2023-correct}. For instance, the Czech sentences``Jan chodí do práce'' (‘Jan goes to work.’) and ``Jan nechodí do práce.''(‘Jan does not go to work.’) are likely to be assigned a high similarity score, despite conveying entirely opposite meanings.

Following \cite{liu-etal-2023-g}, we implement an \textbf{LLM-as-a-judge} approach to evaluate similarity which is described in more detail in Section~\ref{ch:llm_judge}. Because LLMs capture broader semantic and contextual nuances, this approach has the potential to alleviate the limitations observed in earlier metrics such as the negation problem.  
\vspace{-1.25em}
\subsection{Claim Extraction}
\vspace{-0.25em}
Currently, most of the fact-checking methods and benchmarks rely on sentence-level claim extraction, often from artificially constructed sentences. This makes the extraction task trivial. Document-level claim extraction, on the other hand, has been addressed scarcely.

\cite{deng-etal-2024-document} focuses on document-level claim extraction, but in the end, they provide only one ``central claim'' per document as output, which they compare to the reference one generated by a professional fact checker.

\cite{metropolitansky-larson-2025-towards} introduce a document-level claim extraction and evaluation framework. They assess claims across three key attributes: \emph{entailment}, \emph{coverage}, and \emph{decontextualization}. Their approach proposes a multi-stage evaluation network, rigorously designed to assess claim quality for use in decompose-then-verify fact-checking pipelines. Notably, in contrast to other works, they do not incorporate the atomicity and checkworthiness of claims, considering the former as lacking a well-defined endpoint and the latter as too
subjective for reliable automated evaluation.

We propose a simple evaluation framework that captures exhaustivity, penalizing both missing and redundant claims.
This setup allows us to precisely examine how different metrics—specifically LLM-as-a-judge, BERTScore, and Edit similarity—perform and how their results diverge when compared to human judgment. We conduct this analysis on two novel datasets based on Czech and Slovak online news comments, which present unique linguistic challenges due to their informal and diverse nature.
\vspace{-0.5em}
\section{Evaluation Method}
\vspace{-0.75em}
\subsection{Datasets}
\vspace{-0.25em}
The properties of our datasets used for evaluation are summarized in Table~\ref{tab1}. The comments used for their creation are sourced from articles in multiple Czech and Slovak online outlets between 2021 and 2024, and are included in a Hugging Face dataset available at \url{https://huggingface.co/datasets/FactDeMice/CzechNews}.
\subsubsection{Czech Gold Claims Dataset (CGCD)} consists of 60 comments and their corresponding gold-standard claims, which were manually extracted by the first author. Each comment contains on average two claims, and the dataset is entirely in Czech.

\subsubsection{Czech-Slovak Agreement Dataset (CSAD)} includes 120 comments with two-way annotations, meaning each comment was independently labeled by two different annotators. These comments contain on average three claims and exhibit more variation in source text length. They were annotated by both Czech and Slovak annotators from Czech and Slovak comments respectively, resulting in a mixture of the two languages. This dataset can serve for comparing human-human (inter-annotator) agreement to human-model agreement.
\vspace{-1.5em}
\begin{table}[ht]
\centering
\caption{Statistics for Both Datasets.}\label{tab1}
\begin{tabular}{|l|c|c|}
\hline
\textbf{Dataset} &  \textbf{CGCD} & \textbf{CSAD}\\
\hline
Number of samples &  60 & 120\\
Average claims per comment &  2.17 & 3.12 \\
Language & Czech & Czech and Slovak\\
Total annotators & 1 & 9\\
\hline
\end{tabular}
\end{table}
\vspace{-2.0em}

\subsection{Model choice}
\vspace{-0.25em}
For claim extraction, we employ several commercial large language models, specifically \textbf{ChatGPT 4o mini}, \textbf{ChatGPT 4.1}, and \textbf{Claude Sonnet 3.7}. Part of the motivation for using commercial systems at this stage is to reserve the stronger open-source models for use in our LLM-based evaluation framework.

The choice of models for the evaluation part is informed by their performance on the Czech language tasks in the BenCzechMark benchmark \cite{fajcik2025benczechmarkczechcentricmultitask}. Based on these results, we select \textbf{Qwen2.5 (32B)} and \textbf{Phi-4} as our open-weight LLM-based evaluators. We additionally aimed to mitigate potential data leakage to commercial models.
\subsection{Claim Extraction Method}
\vspace{-0.25em}
Our claim extraction method inspired by LOKI \cite{li-etal-2025-loki}, extracts
claims in two steps. 
\begin{enumerate}[label=(\roman*)]
    \item We prompt the model to extract all the claims, emphasizing that they should be atomic and decontextualized.
    \item The model judges which of the claims are check-worthy and retrieves them.
\end{enumerate} 
\vspace{-0.5em}
We use Czech translations of the original ``LOKI prompts'' for these tasks, only slightly adjusted to highlight the need to produce claims that are decontextualized.

\subsection{Alignment and Evaluation}
\vspace{-0.25em}
Since we work at the document level, we obtain sets of claims and the evaluation consists of two tasks:
\begin{enumerate}[label=(\roman*)]
    \item Determining the optimal pairing of the claims.
    \item Scoring the pairs by the similarity metric.
\end{enumerate}
\vspace{-0.5em}
We start by constructing a similarity matrix and computing pairwise similarity scores $score(x,y): \mathcal{C}_1 \times \mathcal{C}_2 \rightarrow \mathbb{R}$ between all combinations of candidate and reference claims using the selected metric, as illustrated in Algorithm \ref{a:eval}, The optimal pairing is then determined by applying the Hungarian algorithm. 

\begin{algorithm}[h]
\label{a:eval}

\caption{Similarity Matrix Construction}
\begin{algorithmic}[1]

\Require Claim sets $\mathcal{C}_1$ and $\mathcal{C}_2$

\State $n \gets \max(|\mathcal{C}_1|, |\mathcal{C}_2|)$
\State Initialize $S \in \mathbb{R}^{n \times n}$ with zeros

\For{$i = 1$ to $|\mathcal{C}_1|$}
    \For{$j = 1$ to $|\mathcal{C}_2|$}
        \State $S_{ij} \gets \text{score}(c_i^{(1)}, c_j^{(2)})$
    \EndFor
\EndFor
\end{algorithmic}
\end{algorithm}
\vspace{-2.0em}

\subsection{LLM-as-judge Metric}
\vspace{-0.25em}
\label{ch:llm_judge}
Our \textit{LLM-as-judge} metric is implemented as a logit-based pair evaluation following~\cite{liu-etal-2023-g}. 
Given a claim pair, the model is prompted to output a single token  $t_s \in \{0,1,...,4\}$ representing a similarity score. 
The logits corresponding to these numerical tokens are extracted and normalized using the softmax function to obtain probabilities $p_{t_s}$ for each discrete value $s$. 

By computing the expected value
\[
s = \sum_{i=0}^{4} p_i\, i,
\]
we derive a continuous similarity measure $s \in [0,4]$. 

\section{Experiments}
For clear and consistent comparison across experiments, all results presented in this chapter are normalized to a standardized 0–100 scale.
\vspace{-0.75em}
\subsection{Similarity Evaluation}
\vspace{-0.25em}
We evaluate the similarity between AI-extracted and human-annotated claims. For \textbf{CSAD} dataset, we apply max pooling, selecting the higher similarity score from the two annotators for each sample. 
The results reported in Table~\ref{tab:similarity_comparison} show: 
\begin{enumerate}[label=(\roman*)]
    \item The metrics agree on the best-performing model on both datasets, but differ in how they rank the remaining models.
    \item Models appear to align with humans more consistently than human annotators align with each other.
    \item Edit Similarity heavily underrates human agreement, even ranking it as the lowest. This reflects known limitations of lexical metrics: although human-written claims were often semantically equivalent, some were written in different but closely related languages, lowering lexical overlap. 
\end{enumerate}
\vspace{-0.5em}
Overall, these findings confirm that document-level claim extraction is challenging, and even humans struggle to maintain consistent agreement. To further investigate, we performed a leave-one-annotator-out analysis and found an outlier. A closer inspection revealed, that this annotator misunderstood certain aspects of the task. Excluding his annotations, inter-annotator agreement improved---\textbf{substantially exceeding the best model-to-human performance} (under LLM-as-a-judge: from 47.7 to 51.8 , while GPT-4o-mini’s new score was only 49.8). This suggests that our metric can support annotation quality assessment even without gold-standard labels. 
\vspace{-1.5em}
\begin{table}[ht]
\centering
\caption{Combined results for the Czech Gold Claims (CGCD) and Czech–Slovak Agreement (CSAD) datasets. Models are evaluated using BERTScore, Edit Similarity, and two LLM-as-judge metrics: LLM–P (Phi-4 as judge) and LLM–Q (Qwen 2.4 as judge). Bold values indicate the best score for each metric.}
\label{tab:similarity_comparison}
\begin{tabular}{|l|c|c|c|c|c|}
\hline
\textbf{Data} & \textbf{Model} & \textbf{BERTScore} & \textbf{LLM-P} & \textbf{LLM-Q} & \textbf{Edit Similarity} \\
\hline
 \textbf{CGCD} & Sonnet 3.7    & 52.8 & 41.3 & 40.9 & 35.6 \\
 \hline
 & \textbf{GPT 4o mini} & \textbf{60.2} & \textbf{45.3} & \textbf{43.8} & \textbf{37.9} \\ 
 \hline
 & GPT 4.1       & 54.3 & 43.4 & 42.2 & 36.7 \\
 \hline
 & GPT 4.1 (c)   & 54.7 & 41.8 & 43.4 & 35.8 \\
\hline\hline
\textbf{CSAD} & a1 vs a2      & 53.8 & 47.7 & -- & 28.3 \\
 \hline
 & Sonnet 3.5    & 53.9 & 44.7 & -- & 30.7 \\
 \hline
 & \textbf{GPT 4o mini} & \textbf{58.3} & \textbf{49} & -- & \textbf{34.9} \\
 \hline
 & GPT 4.1       & 55.2 & 46 & -- & 32.7 \\
 \hline
 & GPT 4.1 (c)   & 52.6 & 44.7 & -- & 31.2 \\
 \hline
\end{tabular}
\end{table}
\vspace{-2.5em}

\subsection{Claim Count Differences}
\vspace{-0.25em}
We distinguish two core abilities evaluated by our framework:
\begin{enumerate}[label=(\roman*)]
    \item extracting the correct number of check-worthy claims
    \item extracting claims that are semantically similar to the gold ones
\end{enumerate}
\vspace{-0.5em}
First, we assess the quantity aspect by measuring the average difference between the number of extracted and annotated claims (Table~\ref{tab:claim_counts_comparison}). 

We also report CGCD informed, which represents the average difference in claim counts when models are explicitly informed of the expected number of claims to extract. All evaluated models became significantly more accurate in matching the expected number of claims on average under this informed setting.
The similarity results for this setup are presented in the next subsection.


For the \textbf{CSAD} dataset, we evaluate annotator agreement by measuring the average absolute difference between their claim-set sizes. We provide two scores, both representing differences: \textbf{CSAD-min}, which is the minimum absolute difference between the number of model-extracted claims and the claim-set size of either annotator (minimum taken across the two annotators), and \textbf{CSAD-total}, which is the absolute difference between the number of model-extracted claims and the mean claim-set size across both annotators.
\vspace{-1.5em}
\begin{table}[ht]
\centering
\caption{Average claim-count difference between model outputs and reference annotations. GPT-4.1 tends to over-generate the most when uninformed, while GPT-4o mini shows the smallest improvement when given the expected claim count.}
\label{tab:claim_counts_comparison}
\begin{tabular}{|l|c|c|c|c|}
\hline
\textbf{Model} & \textbf{CGCD} &\textbf{CGCD informed} & \textbf{CSAD-min} & \textbf{CSAD-total} \\ 
\hline
Human-to-Human               & - & - & 2.07 & 2.07\\
\hline
Sonnet 3.7             & +1.23 & -0.07 & +0.46 & +1.49  \\
\hline
GPT 4o mini            & +0.43 & -0.36 & -0.25 & +0.78 \\
\hline 
GPT 4.1                & +1.65 & -0.1 & +0.9 & +1.93 \\
\hline
GPT 4.1 (c)            & +1.52 & -0.03 & +1.33 & +2.36 \\
\hline
\end{tabular}
\end{table}

\subsection{Effect of Disclosing the Correct Claim Count}
\vspace{-0.25em}
In this controlled setup, models were informed of the expected number of claims before generation. This reduces exhaustivity-related penalties and allows us to more directly evaluate semantic extraction performance. As shown in Table~\ref{tab:similarity_comparison_informed}, this guidance leads to noticeable improvements in similarity scores across all evaluation metrics.

This setup also changes model rankings. Some models that appeared strong in the unconstrained setting benefited mainly from conservative output behavior—producing fewer claims and therefore receiving fewer penalties—rather than superior extraction capability. These findings suggest that refining the penalty component of our method could lead to a more accurate and fair comparison of model performance.
\vspace{-1.5em}
\begin{table}[ht]
\centering
\caption{\textbf{Czech Gold Claims Dataset}: Model comparison across different similarity metrics after informing the model about target number of check-worthy claims.}
\label{tab:similarity_comparison_informed}
\begin{tabular}{|l|c|c|c|c|}
\hline
\textbf{Model} & \textbf{BERTScore} & \textbf{LLM-P} & \textbf{LLM Judge - Q} & \textbf{Edit Similarity} \\
\hline
Sonnet 3.7   & 83.4 & \textbf{66.9} &	\textbf{64}	& \textbf{55.8} \\
\hline
GPT 4o mini  &	66.4	&	51	&	49.5	&	42.2 \\
\hline
GPT 4.1      &	82.8	&	64.6	&	61	&	52 \\
\hline
GPT 4.1 (c)  &	\textbf{84}	&	64.5	&	62.8	&	52.5 \\
\hline
\end{tabular}
\end{table}
\vspace{-2.0em}

\subsection{Alignment with Human Judgement}
\vspace{-0.25em}
To assess how well the metrics align with human perception of extraction quality, we performed a human preference evaluation. Annotators were shown two randomly selected LLM-generated claim sets (from the specified group of LLM extractions) for each comment, along with the source comment and true claim set. The preference was collected as a vote between the two claim sets, with the option to select both when no clear preference was possible. In Table ~\ref{tab:human_judgement} we report results for each of the groups. 
We report results for our four extraction models, divided into an informed group and a non-informed group, and evaluated separately. In each case, we collected 2 annotations for each sample in the dataset, yielding 120 preference judgments per group.

For a more fine-grained comparison, we additionally performed pairwise preference evaluations for the informed versions of Sonnet 3.7 vs. GPT-4.1 and Sonnet 3.7 vs. GPT-4.1 (with context). In the uninformed setting, rankings differ substantially from metric-based rankings, likely because human preference emphasizes clarity and correctness more than strict exhaustivity. In an informed setting, the results align more closely with the metric scores in \ref{tab:similarity_comparison_informed}—particularly with the LLM-as-a-judge metric. The pairwise results additionally support Sonnet 3.7 as the strongest performer in the informed setup.

Overall, these findings indicate that the LLM-as-a-judge metric correlates reasonably well with human judgment, especially when quantity-based penalties are minimized. Refining the penalty component remains important to better reflect actual extraction quality.
\vspace{-1.5em}
\begin{table}[ht]
\centering
\caption{\textbf{Evaluation of human judgements}: Model comparison by human preference judgement. Pair 1 and Pair 2 represent pairwise preference evaluations for the informed versions of Sonnet 3.7 vs. GPT-4.1 and Sonnet 3.7 vs. GPT-4.1 (with context).}
\label{tab:human_judgement}
\begin{tabular}{|l|c|c|c|c|}
\hline
\textbf{Model} & \textbf{Not  informed} & \textbf{Informed} & \textbf{Pair 1} & \textbf{Pair 2} \\
\hline
Sonnet 3.7   & 40 & 53 & 50	& 52 \\
\hline
GPT 4o mini  &	38 & 20 & - & -	 \\
\hline
GPT 4.1      &	28	&	51	& -	&	41 \\
\hline
GPT 4.1 (c)  &	40	&	41	& 39	&	- \\
\hline
\end{tabular}
\end{table}
\vspace{-2.0em}

\section{Conclusion}
\vspace{-0.75em}
We focused on exploring the options for claim extraction at the document level and the evaluation of such methods. We created a small, human-annotated dataset for claim extraction and an additional two-way annotation dataset, the latter posing a greater challenge for extraction models. Using these datasets, we compared multiple claim similarity metrics within our evaluation framework, which addresses both metric choice and the alignment strategy between model outputs and human annotations. The framework also penalizes over- and under-generation.

Our results indicate that 
\vspace{-0.5em}
\begin{enumerate}[label=(\roman*)]
    \item \textbf{The penalization component requires careful tuning}: reducing its impact improves alignment with human judgment and highlights the need for better calibration to produce the appropriate number of claims.
    \item \textbf{Claim extraction from real-world sources remains challenging}: models perform substantially better when given the correct number of expected claims, suggesting that estimating this number is a key bottleneck.
    \item \textbf{Our LLM-based evaluation metric shows promise}: it mitigates several limitations of existing approaches while demonstrating strong correlation with human preferences.
\end{enumerate}
\vspace{-0.75em}
\section{Limitations and Future Work}
\vspace{-0.75em}
Our LLM-as-judge metric does not yet explicitly evaluate checkworthiness or decontextualization. Currently, these properties are enforced only indirectly through high-quality reference claims. Incorporating them directly into the metric would allow model comparison without gold-standard annotations, enabling practical preference-based evaluation (e.g., tournament-style setups like BenCzechMark). 
We also haven't given importance to another attribute essential for fact-checking -disambiguation.
Lastly, given the small size of our datasets, these findings should be further validated on a larger scale.

\subsubsection{Acknowledgements.} This work was supported by the Technology Agency of the Czech Republic (TAČR) under the SIGMA Programme, 8th Public Competition, Sub-objective 4: Bilateral Cooperation, project TQ16000028.
%
%
%
\bibliographystyle{splncs04}
\bibliography{refs}
\end{document}